\documentclass[a4paper]{article}

\usepackage{INTERSPEECH2022}
\usepackage{multirow,multicol}
\usepackage{bigstrut}
\usepackage{comment}
\usepackage{color}
\usepackage{makecell}
\usepackage{caption}
\usepackage[hyphens]{url}
\usepackage{hyperref}
\usepackage[hyphenbreaks]{breakurl}
\title{Open Source MagicData-RAMC: A Rich Annotated Mandarin Conversational(RAMC) Speech Dataset}
\name{{Zehui Yang$^{1,2,*}$, Yifan Chen$^{1,2,*}$}\thanks{$^*$ Equal contribution.}, Lei Luo$^{3,\dagger}$\thanks{$^\dagger$ Corresponding author.}, Runyan Yang$^{1,2}$, Lingxuan Ye$^{1,2}$, Gaofeng Cheng$^{1}$,Ji Xu$^{1}$, Yaohui Jin$^4$, Qingqing Zhang$^3$, Pengyuan Zhang$^{1,2}$, Lei Xie$^5$, Yonghong Yan$^{1,2}$}
\address{
  $^1$Key Laboratory of Speech Acoustics and Content Understanding, Institute of Acoustics, Chinese Academy of Sciences, China \\
  $^2$University of Chinese Academy of Sciences, China\\
  $^3$Magic Data Technology Co., Ltd., China\\
  $^4$MoE Key Lab of Artificial Intelligence and AI Institute, Shanghai Jiao Tong University, China\\
  $^5$Northwestern Polytechnical University, China}
\email{chenggaofeng@hccl.ioa.ac.cn, luolei@magicdatatech.com, zhangqingqing@magicdatatech.com}
\begin{document}

\maketitle
\begin{abstract}
This paper introduces a high-quality rich annotated Mandarin conversational (RAMC) speech dataset called MagicData-RAMC. The MagicData-RAMC corpus contains 180 hours of conversational speech data recorded from native speakers of Mandarin Chinese over mobile phones with a sampling rate of 16 kHz. The dialogs in MagicData-RAMC are classified into 15 diversified domains and tagged with topic labels, ranging from science and technology to ordinary life. Accurate transcription and precise speaker voice activity timestamps are manually labeled for each sample. Speakers' detailed information is also provided. As a Mandarin speech dataset designed for dialog scenarios with high quality and rich annotations, MagicData-RAMC enriches the data diversity in the Mandarin speech community and allows extensive research on a series of speech-related tasks, including automatic speech recognition, speaker diarization, topic detection, keyword search, text-to-speech, etc. We also conduct several relevant tasks and provide experimental results to help evaluate the dataset.

% MagicData-RAMC aims to fill a gap in Mandarin speech database and benefit the exploration of human-human and human-computer interaction. With dialog speech in spontaneous speaking style, MagicData-RAMC presents realistic speech characteristics in real world dialogs 
% The dialogs in the dataset reflect our daily communication way and cover various topics about our daily life.

  %This paper considers dialog scenario and contributes a high quality Chinese speech dataset. The dataset contains 160 hours conversation data from 587 speakers. It also provides voice acticitiy timestamps of each speaker. 
% 	Besides, baselines of speech recognition task and speaker diarization task are provided to evaluate the dataset.

\end{abstract}

\section{Introduction}

As an essential resource for the speech research community, speech data is especially required for spoken language processing technologies based on statistical models \cite{graves2014towards, chan2016listen, 8068205, Dong2018SpeechTransformerAN, gulati2020conformer,9072325} which are data-driven. Over the past few decades, a growing number of well-curated and freely available speech corpora with various accents, speaking styles, recording environments, channels, scales, etc., have been developed and released, trying to cover the infinite diversity of human speech. 

% Different scales of English datasets cover various scenarios, ranging from read speech, broadcast speech, telephony speech, meeting speech to speech collected from YouTube and podcasts, etc. 
There have been many datasets created for general speech recognition tasks. The Wall Street Journal corpus \cite{paul92_icslp} is a classic dataset including 80 hours of narrated news articles. LibriSpeech \cite{panayotov2015librispeech}, one of the most widely used English corpora, is a reading speech dataset based on LibriVox's audiobooks with steady speed and consistent tone. TED-LIUM 3 \cite{hernandez2018ted} consists of 452 hours of audio collected from TED talks. THCHS30 \cite{THCHS30_2015} is a toy Chinese speech database accessible to academic users. AISHELL-1 \cite{bu2017aishell} is a commonly used Mandarin dataset with speeches in reading style. These datasets are relatively formal in language and recorded in a single-speaker manner; hence lack natural speech characteristics and interaction between speakers. WenetSpeech \cite{zhang2021wenetspeech} is a recently released Mandarin corpus containing more than 10000 hours of speech collected from YouTube and podcasts, adopted with optical character recognition (OCR) and automatic speech recognition (ASR) methods to generate the transcriptions, respectively. GigaSpeech \cite{Chen2021GigaSpeechAE} is a large-scale English corpus with audio collected from audiobooks, podcasts, and YouTube. Its transcripts are downloaded from various sources and applied with standard text normalization, and the word error rate of the training subset is around $4\%$.

%  limits their application and generalization
% LibriCSS \cite{Chen2020ContinuousSS} is a real recording dataset derived from LibriSpeech. By replaying the mixture of the utterances from different speakers in LibriSpeech and recording the audio by a microphone array in a conference venue, the corpus attempts to simulate natural conversations. 
To expand application areas and conduct more tasks such as speaker diarization, some corpora are collected among multi-speakers in more diverse scenarios. Meeting speech datasets, such as ISCI \cite{janin2003icsi}, CHIL \cite{mostefa2007chil} and AMI \cite{ami}, are composed of audio recorded during the conferences held in academic laboratories. AISHELL-4 \cite{fu2021aishell} is a 120-hour Mandarin speech dataset collected in conference scenarios. The number of attendees in each session is between four and eight. The indoor conversation corpus CHiME-6 \cite{watanabe2020chime} consists of unsegmented recordings of twenty separate dinner-party conversations among four friends captured by Kinect devices. It has a more casual and natural speaking style yet relatively low recording quality. For dialog scenario, Switchboard \cite{swbd} and Fisher \cite{fisher} are classic datasets of English telephony conversations with similar settings and different scales. HKUST is a Mandarin conversational corpus \cite{liu2006hkust} made up of spontaneous telephone conversations. Audios in these three telephony conversational datasets are recorded with a sampling rate of 8 kHz, which is incompatible with the demand of some speech processing systems nowadays. 

% Compared with English corpora, Mandarin Chinese speech community suffers from lack of data variations more severely despite the popularity of Chinese. 

% While all of these datasets facilitate the research in speech-relevant fields, the applications based on it sometimes fail in the complex real world. 
% Compared with English corpora, Mandarin Chinese speech community suffers from lack of data variations more severely despite the popularity of Chinese. 

Although various corpora, including the ones mentioned above, have been introduced, most of them are not specifically designed for dialog scenarios. Meanwhile, the utilization of the few existing dialog corpora is limited by collection setups such as sampling rate. To the best of our knowledge, for Mandarin Chinese, there is no public-available dialog speech dataset adequate for the current requirement of high quality.
% Since there has been growing research interest in dialog speech recognition and modelling of interactions between people
With the boom in popularity of voice-driven interfaces to devices recently, some works \cite{kim2018dialog, 9414575} concerned with communication scenes have been conducted. However, exploring speech processing techniques in dialog scenarios is still challenging. Deficient and mismatched data is one of the limitations that constrict the investigation of communication scenarios due to the attributes of dialog speech.

In order to enrich the diversity of the Mandarin database and alleviate the scarcity of data specialized in dialog scenarios, we develop a high-quality rich annotated Mandarin conversational corpus in this work and refer to it as MagicData-RAMC\footnote{\url{https://www.magicdatatech.com/datasets/mdt2021s003-1647827542}. The partition of the dataset and the keyword list used in the subsequent KWS experiment are provided together in the URL.}. It contains 180 hours of dialog speech recorded over mobile phones with a sampling rate of 16 kHz. Accurate transcriptions are manually labeled and proofed for each sample. Precise voice activity timestamps of each speaker, each sample's topic label, and speakers' demographic information are also provided, allowing further research on different tasks. The dialogs in MagicData-RAMC aim to reflect our realistic communication in the real world and cover various topics about our daily lives. On the one hand, there are rich and complicated speech characteristics in its samples, including colloquial expressions, hesitations, repetitions, nonsense syllables, and other speech disfluencies. On the other hand, during long dialogs in the dataset, people respond to each other in a flexible manner and continue the dialog with coherent questions and opinions instead of stiffly answering each other's questions. Therefore, history and current utterances are closely related, and a consistent topic runs through the conversation by the contextual flow. We believe that MagicData-RAMC can facilitate the research on a series of tasks relevant to multi-turn dialogs, and we conduct several of them to evaluate the dataset.
 
% Since speech is considered to be a natural way for humans to communicate, conversational scenarios are common and critical for human speech, where large amounts of information, ideas and emotion can be exchanged.
% Relevant researches are fundamental to high-level applications, such as voice assistants and personalized chatbots, and developing intelligent conversational agents is of great potential and significance in both academic and commercial fields. 

The rest of this paper is organized as follows. We introduce the construction process of MagicData-RAMC and present its structure and details in Section~\ref{sec:2}. Then we describe the baseline systems built on popular toolkits for speech recognition, speaker diarization (SD), and keyword search (KWS) tasks and provide experiment setup and results in Section~\ref{sec:3}. Finally, we conclude the entire paper in Section~\ref{sec:4}.

% In recent times there has been growing research interest in the recognition and understanding of interactions between people in settings such as meetings, lectures, seminars and teleconferences. The analysis and interpretation of multiparty meetings is of scientific interest since it provides a circumscribed arena for the investigation of communication scenes, as well as underpinning a number of potentially significant applications.

% With speech recognition technology extensively studied in recent years, speech tasks such as signal processing, automatic speech recognition (ASR) and speaker diarization (SD) are arising more and more attention. 

\section{Dataset Description}
\label{sec:2}
% In this paper, a new high quality Chinese dialog dataset named "MAGICDATA-DIALOG" is presented. 
% This dataset contains 160 hours Chinese conversation data. The sampling rate is 16k. The  annotated For 
% During dataset collection

MagicData-RAMC comprises dialog speech data, corresponding transcriptions, voice activity timestamps, and speakers' demographic information. It contains 351 multi-turn Mandarin Chinese dialogs, which amount to about 180 hours. The speech data is carefully annotated and manually proofed.

\subsection{Collection setup}
The dataset is collected indoors. The domestic environments are small rooms under 20 square meters in area, and the reverberation time (RT60) is less than 0.4 seconds. The environments are relatively quiet during recording, with ambient noise level lower than 40 dBA. 

The audios are recorded via an application developed by Magic Data Technology Co., Ltd.\footnote{\url{https://www.magicdatatech.com}} over mainstream smartphones, including Android phones and iPhones. The ratio of Android phones to iPhones is around $1: 1$. All recording devices work at 16 kHz, 16-bit to guarantee high recording quality.

All speech data are manually labeled using the platform built by Magic Data Technology Co., Ltd. Sound segments without semantic information during the conversations, including laughter, music, and other noise events, are annotated with specific symbols. Phenomena common in spontaneous communications, such as colloquial expressions, partial words, repetitions, and other speech disfluencies, are recorded and fully transcribed. Punctuation is carefully checked to ensure accuracy. We also segment the dialog speech and provide precise voice activity timestamps of each speaker. The transcriptions are proofed by professional inspectors to ensure the labeling and segmentation quality. 

\subsection{Speaker information}

There are a total of 663 speakers involved in the recording, of which 295 are female and 368 are male. Each segment is labeled with the corresponding speaker-id. All participants are native and fluent Mandarin Chinese speakers with slight variations of accent. The accent region is roughly balanced, with 334 Northern Chinese speakers and 329 Southern Chinese speakers. The detailed distribution based on the birthplaces of the speakers is shown in Table~\ref{tab:spk}. Besides, each speaker participates in up to three conversations.

\begin{table}[th]
  \caption{Region distribution of speakers}
  \label{tab:spk}
  \centering
  \begin{tabular}{c c c c}
    \toprule
    \textbf{Region} & \textbf{Province} & \textbf{\#Speaker} & \textbf{Total}  \\
    \midrule
    \multirow{3}{*}{Northern China} & Shandong & $94$  &  \multirow{3}{*}{334}     \\
     & Shanxi  & $228$   &            \\
     & Beijing  & $12$   &            \\
     \midrule
    \multirow{4}{*}{Southern China}  & Guangdong  & $12$  &  \multirow{4}{*}{329}      \\
     & Hunan  & $249$     &            \\
     & Jiangsu & $10$      &            \\
     & Sichuan  & $58$    &           \\
    \bottomrule
  \end{tabular}
  \vspace{-0.1in}
\end{table}

\subsection{Dataset Partition}
We divide the corpus into 150 hours training set, 10 hours development set, and 20 hours test set, containing 289, 19, and 43 conversations, respectively. The partition of the speech data is provided in TSV format. There are 556, 38, and 86 speakers split into three subsets. The gender and region distribution is roughly proportional to the entire dataset. Table~\ref{tab:partition} provides a summary of the partition of the corpus.
\begin{table}[htbp]
  \caption{Corpus partition}
  \label{tab:partition}
  \centering
  \begin{tabular}{l c c c}
    \toprule
     & {\textbf{Training}} & \textbf{Development} &\textbf{Test}\\
    \midrule
    Duration (h) & $149.65$ & $9.89$    & $20.64$         \\
    \#Sample & $289$  & $19$  & $43$             \\
    \#Speaker  & $556$  & $38$  &  $86$   \\
    \#Male  & $307$  & $23$    &  $49$ \\
    \#Female & $249$  & $15$    &   $37$   \\
    \#Northern & $271$ & $20$ & $52$\\
    \#Southern & $285$ & $18$ & $34$\\
    \bottomrule
  \end{tabular}
  \vspace{-0.1in}
\end{table}

\subsection{Utterance statistics}

The whole dataset is composed of 351 conversations with 219325 segments in total. Each conversation is of 30.80 minutes duration and segmented to about 625 speech segments on average. The start and end times of all segments are specified to within a few milliseconds. We also count the length of segments and the number of tokens per segment to give a simple and intuitive view of the corpus. The statistics are presented in Table~\ref{tab:speech}.

\begin{table}[th]
  \caption{Statistics of speech information}
  \label{tab:speech}
  \centering
  \begin{tabular}{l c c c}
    \toprule
    \textbf{Statistical Criterion} & {\textbf{Max}} & {\textbf{Min}} & {\textbf{Average}} \\
    % \textbf{Statistical} & \multirow{2}{*}{\textbf{Max}} & \multirow{2}{*}{\textbf{Min}} & \multirow{2}{*}{\textbf{Average}} \\
    %  \textbf{Criterion} & & & \\
    \midrule
    Sample Duration (min) & $33.02$ & $14.06$  &  $30.80$     \\
    \#Segments Per Sample & $1215$  & $231$   &    $624.86$        \\
    Segment Duration (s) & $14.91$  & $0.09$   &  $2.54$      \\
    \#Tokens Per Segment & $89$ & $1$ & $13.58$\\
    \#Segments Per Speaker & $1155$ & $46$ & $304.55$ \\
    \bottomrule
  \end{tabular}
  \vspace{-0.1in}
\end{table}

The dialogs in MagicData-RAMC aim to reflect the natural communication way in the real world and cover a broad range of topics closely relevant to our daily life. During the multi-turn conversations in the dataset, people respond flexibly to each other and continue the dialog with relevant questions and items instead of replying and waiting for the following questions rigidly. Therefore, every sample is a coherent and compact conversation centered around one theme, with history utterances and current utterance closely related. Higher-level information is maintained by contextual dialog flow across multiple sentences. 

We classify the dialogs in MagicData-RAMC into 15 diversified domains, ranging from science and technology to ordinary life, and provide topic labels for them. The diversity of topics and the consistency in one dialog are beneficial to the development of open-domain spoken dialog systems. We summarize the statistics of the categories in Table~\ref{tab:topic}.

\begin{table}[htbp]
  \caption{The distribution over topics}
  \label{tab:topic}
  \centering
  \begin{tabular}{l c c}
    \toprule
    {\textbf{Topic}} & {\textbf{\#Sample}} & \textbf{Duration (h)}\\
    \midrule
    Humanities & $22$ & $11.46$            \\
    Entertainment & $1$  & $0.52$               \\
    Sports  & $32$  & $16.62$       \\
    Military  & $4$  & $1.98$             \\
    Finance \& Economy & $5$  & $2.49$              \\
    Religion & $1$ & $0.52$     \\
    Family Life & $6$ & $1.48$   \\
    Politics \& Law  & $4$  & $2.07$              \\
    Education \& Health  & $53$  & $27.30$              \\
    Digital Devices  & $39$ & $20.14$ \\
    Climate \& Environment & $13$ &  $6.62$ \\
    Science \& Technology& $11$ &  $5.66$ \\
    Professional Development & $35$ &  $18.26$ \\
    Art & $84$ &  $43.87$ \\
    Ordinary Life& $41$ &  $21.19$ \\
    \bottomrule
  \end{tabular}
  \vspace{-0.1in}
\end{table}

\subsection{Comparison with other datasets}

The main properties of several public-available manually labeled speech datasets are summarized in Table~\ref{tab:compare} for a brief comparison. Compared to other corpora, MagicData-RAMC is the most suitable conversational speech dataset for research on Mandarin dialog speech. 

\begin{table*}[htbp]
    \setlength\tabcolsep{5pt}
	\caption{Comparison with several manually labelled datasets}
	\label{tab:compare}
	\begin{center}
		\newcommand{\tabincell}[2]{\begin{tabular}{@{}#1@{}}#2\end{tabular}}
% 		 \resizebox{\textwidth}{20mm}{
		\begin{tabular}{clcccccccc}
			\toprule
% 			\multirow{2}{*}{\textbf{Language}} & \multirow{2}{*}{\textbf{Datasets}}  &   \textbf{Duration} & \multirow{2}{*}{\textbf{Timestamps}} & \multicolumn{3}{c}{\textbf{Spk Info}}  & \multirow{2}{*}{\textbf{Device}} & \textbf{Sampling}\bigstrut[t]\\
	         \multirow{2}{*}{\textbf{Language}} & \multirow{2}{*}{\textbf{Datasets}}  &   \textbf{Duration} & \multirow{2}{*}{\textbf{Scenario}} & \multirow{2}{*}{\textbf{Topic label}} &\multirow{2}{*}{\textbf{\#Spkrs}} & \multirow{2}{*}{\textbf{Device}} & \textbf{Sampling}\bigstrut[t]\\			
% 			\cline{5-6}
			&  &  \textbf{(hours)} &  &  &   & &\textbf{Rate (kHz)}\bigstrut\\
			\midrule
			\multirow{3}{*}{EN} & LibriSpeech\cite{panayotov2015librispeech} 	& 1000 &	Reading& 	& 2484 	& mic & 16\\
			&AMI\cite{ami} &  100 &	Conference &  \checkmark	& 200 & mic array, headsets & 16\\
			&Switchboard\cite{swbd} & 317 &	Convensation & 	& 500 & telephone & 8\\
			\midrule
			\multirow{4}{*}{CN}&THCHS-30\cite{THCHS30_2015}  & 33 &Reading	& &  40	& mic & 16 \\
			&MAGICDATA-READ & 755 	& 	Reading	& \checkmark	& 1080 		& mobile phone, mic& 16\\
% 			&AISHELL-4\cite{fu2021aishell} & 120 & \checkmark & 4-8 	& 61  & mic array& 16 \bigstrut[t]\\
			&HKUST\cite{liu2006hkust} & 200 & Convensation	& \checkmark 	& 2412	 	& telephone & 8 \\
			&\textbf{MagicData-RAMC}& 180 	& \textbf{Convensation}	& \textbf{\checkmark}	& 663 	& mobile phone & \textbf{16}\bigstrut[b]\\
			\bottomrule
		\end{tabular} %}
	\end{center}
	\vspace{-0.1in}

\end{table*}
% \footnotetext[2]{http://www.openslr.org/68/}

The quality of MagicData-RAMC is comparable to open Mandarin speech datasets collected in a single-speaker manner. At the same time, MagicData-RAMC is closer to applications in the real world due to its natural spontaneous speaking style and realistic speech characteristics as a corpus specifically designed for dialog scenarios.
% Although conference is another typical speech interactive scenario, it is quite different from daily communication due to its formality and number of attendees. 
MagicData-RAMC is also different from conference datasets in the number of attendees, though they are all developed in speech interactive scenarios. 
Compared to classic telephony conversational datasets with a sampling rate of 8 kHz, the audios recorded via mobile phones and sampled at 16 kHz in MagicData-RAMC are more compatible with most of the current speech processing systems' higher requirements for recording quality. What's more, rich annotations and detailed information in MagicData-RAMC allow further exploration of various speech processing tasks.

\section{Experiments}
\label{sec:3}
In this section, we build baseline systems for ASR, SD, and KWS tasks and present the experimental setup and results, respectively, to evaluate this dataset.
%\footnote{\url{https://github.com/MagicHub-io/Magic-Data-ASR-SD-Challenge}} 

\subsection{Automatic Speech Recognition}
\label{sec:asr}

We use a Conformer-based end-to-end (E2E) model implemented by ESPnet2 toolkit \cite{watanabe2018espnet} to conduct speech recognition. The Conformer model is composed of a Conformer encoder proposed in \cite{Gulati2020ConformerCT} and a Transformer decoder. We adopt 14 Conformer blocks in the encoder and 6 Transformer blocks in the decoder. Connectionist temporal classification (CTC) \cite{CTC} is employed on top of the encoder as an auxiliary task to perform joint CTC-attention (CTC/Att) training and decoding \cite{hybridCTCAtt} within the multi-task learning framework. During the beam search decoding, we set the beam size to 10.

We compute 80-dimensional logarithmic filterbank features from the raw waveform and utilize utterance-wise mean variance normalization. The frame length is 25 ms with a stride of 10 ms. SpecAugment \cite{Park2019SpecAugmentAS} is applied with 2 frequency masks and 2 time masks for data augmentation. The maximum widths of each frequency mask and time mask are $F = 30$ and $T = 40$ respectively. The input sequence is sub-sampled through a 2D convolutional layer by a factor of 4. The inner dimension of position-wise feed-forward networks in both encoder and decoder is 2048. We apply dropout in the encoder layer with a rate of 0.1 and set the label smoothing weight to 0.1 for regularization. The multi-head attention layer contains 8 heads with 256 attention transformation dimensions. CTC weight used for multi-task training is 0.3. We train the model using the Adam optimizer with $\beta_1 = 0.9$, $\beta_2 = 0.98$, $\epsilon = 10^{-9}$, gradient clipping norm 5 and no weight decay. Noam learning rate scheduler is set to 25000 warm-up steps. The maximum trainable epoch is 100. The final model is averaged by the last 5 checkpoints.

The training set is made up of two parts: the 150 hours training set of MagicData-RAMC and the 755 hours MAGICDATA Mandarin Chinese Read Speech Corpus (MAGICDATA-READ\footnote{\url{http://www.openslr.org/68}}). The two sets are combined to compose over 900 hours of data for training. 
%The 10 hours development set of the MagicData-RAMC are reserved for validation. 
The experimental result is shown in terms of character error rate (CER) in Table~\ref{asr_result}.

\begin{table}[htbp]
    \caption{ASR Results (CER\%) on dev and test set}
	\label{asr_result}
% 	\begin{center}
  \centering
		\begin{tabular}{ccc}
			\toprule
			Methods & Dev & Test \bigstrut\\
			\hline
			LAS-Conformer & 16.5 & 19.1 \bigstrut[t]\\
% 			LAS-Conformer & 16.5 \bigstrut[b]\\
			\bottomrule
		\end{tabular}
		\vspace{-0.1in}
% 	\end{center}
\end{table}

\subsection{Speaker Diarization}

For SD task, our baseline system consists of three components: speaker activity detection (SAD), speaker embedding extractor and clustering. 

Following Variational Bayes HMM x-vectors (VBx)~\cite{landini2022bayesian} experiment setting, Kaldi-based SAD module \cite{povey2011kaldi} is used for detecting speech activity. We adopt ResNet \cite{he2016deep} trained on VoxCeleb Dataset \cite{DBLP:journals/corr/NagraniCZ17} (openslr-49\footnote{\url{http://www.openslr.org/49}}), CN-Celeb Corpus \cite{fan2020cn} (openslr-82\footnote{\url{http://www.openslr.org/82}}) and the split training set of MagicData-RAMC to obtain the speaker embedding extractor. 

For training details, the SAD module utilizes a 40-dimensional Mel frequency cepstral coefficient (MFCC) with 25 ms frame length and 10 ms stride as input features to detect the speech activity. ResNet-101 with two fully connected layers is employed to conduct speaker classification task with 64-dimensional filterbank features extracted every 10 ms with 25 ms window, and additive margin softmax \cite{wang2018additive} is used to get a more distinct decision boundary. The raw waveform is split every 4s (400 dimensions) to form ResNet input. We train the speaker embedding network using stochastic gradient descent (SGD) optimizer with a $0.9$ momentum factor and $0.0001$ L2 regularization factor. 

Besides, 256-dimension embeddings are conducted dimensionality reduction using probabilistic linear discriminant analysis (PLDA) \cite{ioffe2006probabilistic} to 128-dimension. Embeddings are extracted on SAD result every 240 ms, and the chunk length is set to 1.5s. For the clustering part, we use Variational Bayes HMM~\cite{landini2022bayesian} on this task. An agglomerative hierarchical clustering algorithm with VBx is conducted to get the clustering result. In the VBx, the acoustic scaling factor $Fa$ is set to $0.3$, and the speaker regularization coefficient is set to $17$. The probability of not switching speakers between frames is $0.99$. We present the experimental result in Table~\ref{sd_result}.

\begin{table}[htbp]
    \caption{Speaker diarization results of VBx system}
	\label{sd_result}
	\begin{center}
		\begin{tabular}{ccccc}
			\toprule
			\multirow{2}{*}{Method} & \multirow{2}{*}{Subset} & \multicolumn{2}{c}{DER} & \multirow{2}{*}{JER} \\ % \bigstrut\\
			\cline{3-4}
			 & & collar 0.25 & collar 0 & \bigstrut\\
			\hline
			\multirow{2}{*}{VBx} & Dev & 5.57 & 17.48 & 45.73 \bigstrut\\
			 & Test & 7.96 & 19.90 & 47.49 \\% \bigstrut\\
			\bottomrule
		\end{tabular}
	\vspace{-0.1in}
	\end{center}
\end{table}

\subsection{Keyword Search}

We carry out the KWS task following the DTA Att-E2E-KWS approach proposed in \cite{9576572} relying on attention-based E2E ASR framework and frame-synchronous phoneme alignments. 
The KWS system is based on our Conformer-based E2E ASR system described in Sec~\ref{sec:asr}. 
We adopt the dynamic time alignment (DTA) algorithm to connect a frame-wise phoneme classifier's output posteriors and the label-wise ASR result candidates for generating accurate time alignments and reliable confidence scores of recognized words. 
Keyword occurrences are retrieved within the N-best hypotheses generated in the joint CTC/Att decoding process.
% The reliability is increased by the combination of DTA and attention confidence scores. 
% After getting timestamps and confidence scores for each ASR hypothesis generated in the joint CTC/Att decoding process, we retrieve keywords within the N-best hypotheses. The keywords are firstly retrieved within each of the N-best hypotheses to obtain N sets of keyword occurrences. Then the results are merged to combine the N sets into one set. If a given keyword is found in more than one hypotheses with overlapping timestamp, only the occurrence with the highest confidence score will be kept in the final results.

% The auxiliary CTC branch is exploited to perform joint decoding and provide coarse keyword timestamps for the predictions using CTC time alignment algorithm. We use the attention confidence score obtained by averaging out the decoder's log softmax probabilities of each label in the word. 
 
The keyword list is built by picking 200 words from the dev set and provided together with the dataset.
% lexicon\footnote{\url{https://github.com/aishell-foundation/DaCiDian}}. 
In the DTA Att-E2E-KWS system, the frame-wise phoneme classifier of the KWS system shares 12 Conformer blocks with the E2E ASR encoder while retaining the top 2 Conformer blocks unshared. The classifier outputs posteriors of 61 phonemes, including silence and noise. The KWS system is optimized following the setup in Sec~\ref{sec:asr}. During the inference stage, we retrieve keywords within ASR 2-best hypotheses. During KWS scoring, a predicted keyword occurrence is considered correct when there is a 50\% time overlap at least between the predicted occurrence and a reference occurrence of the same keyword \cite{9576572}. The results are shown in Table~\ref{kws_result}. 

\begin{comment}
关键词表200词
175个：dev中出现次数3~49
25个：dev中出现次数50~299

search within ASR n-best hypothesis: (F1 50%)
        precision   recall      F1 score
n=1     0.8810      0.8814      0.8812
n=2     0.8698      0.8957      0.8826

\begin{table}[htbp]
    \caption{Results on test set for baseline KWS system}
	\label{kws_result}
    \centering
		\begin{tabular}{cccc}
			\toprule
			Methods & Precision rate & Recall rate & F1 score \bigstrut\\
			\hline
			\multirow{2}{*}{\makecell{Conformer DTA \\ Att-E2E-KWS}} & \multirow{2}{*}{0.8698} & \multirow{2}{*}{0.8957} & \multirow{2}{*}{0.8826} \\
			&&&\\
			\multirow{2}{*}{\makecell{Conformer DTA \\ Att-E2E-KWS}} & \multirow{2}{*}{0.8587} & \multirow{2}{*}{0.8879} & \multirow{2}{*}{0.8731} \bigstrut[t]\\
			&&&\\
			\bottomrule
		\end{tabular}
\end{table}
\end{comment}

\begin{table}[htbp]
	\setlength{\belowcaptionskip}{0pt}
    \caption{Results on dev and test set for the Conformer-based DTA Att-E2E-KWS system}
	\label{kws_result}
    \centering
		\begin{tabular}{cccc}
			\toprule
			Subset & Precision rate & Recall rate & F1 score \bigstrut\\
			\hline
			Dev & 0.8698 & 0.8957 & 0.8826 \bigstrut[t] \\
			Test & 0.8587 & 0.8879 & 0.8731 \\ %\bigstrut[t]\\
			\bottomrule
		\end{tabular}
		\vspace{-0.1in}
\end{table}

\section{Conclusions}
\label{sec:4}
In this paper, we release MagicData-RAMC, a high-quality rich annotated Mandarin conversational speech dataset. It is a freely available high-quality Mandarin corpus specially created for dialog scenarios with rich annotations, including precise voice activity timestamps of each speaker, topic labels, etc. We introduce the collection, structure, and detailed analysis of the dataset. We also conduct speech recognition, speaker diarization, and keyword search tasks based on popular speech toolkits to provide examples of the wide utilization of the corpus. We hope that the MagicData-RAMC speech dataset can enrich the diversity of the speech database and facilitate the applications of various speech-related research.

\bibliographystyle{IEEEtran}
\bibliography{mybib}

% \begin{thebibliography}{9}
% \bibitem[1]{Davis80-COP}
%   S.\ B.\ Davis and P.\ Mermelstein,
%   ``Comparison of parametric representation for monosyllabic word recognition in continuously spoken sentences,''
%   \textit{IEEE Transactions on Acoustics, Speech and Signal Processing}, vol.~28, no.~4, pp.~357--366, 1980.
% \bibitem[2]{Rabiner89-ATO}
%   L.\ R.\ Rabiner,
%   ``A tutorial on hidden Markov models and selected applications in speech recognition,''
%   \textit{Proceedings of the IEEE}, vol.~77, no.~2, pp.~257-286, 1989.
% \bibitem[3]{Hastie09-TEO}
%   T.\ Hastie, R.\ Tibshirani, and J.\ Friedman,
%   \textit{The Elements of Statistical Learning -- Data Mining, Inference, and Prediction}.
%   New York: Springer, 2009.
% \bibitem[4]{YourName17-XXX}
%   F.\ Lastname1, F.\ Lastname2, and F.\ Lastname3,
%   ``Title of your INTERSPEECH 2022 publication,''
%   in \textit{Interspeech 2022 -- 23\textsuperscript{rd} Annual Conference of the International Speech Communication Association, September 18-22, Incheon, Korea, Proceedings, Proceedings}, 2022, pp.~100--104.
% \end{thebibliography}

\end{document}